\documentclass{article}

\usepackage{ijcai19}

\usepackage{times}
\usepackage{soul}
\usepackage{url}
\usepackage[hidelinks]{hyperref}
\usepackage[utf8]{inputenc}
\usepackage[small]{caption}
\usepackage{graphicx}
\usepackage{amsmath}
\usepackage{booktabs}
\usepackage{algorithm}
\usepackage{algorithmic}
\usepackage{times}  
\usepackage{helvet}  
\usepackage{courier}  
\usepackage{url}  
\usepackage{graphicx} 
\usepackage{caption}
\usepackage{amsmath}
\usepackage{amssymb}
\usepackage{multirow}
\usepackage{color}
\usepackage{adjustbox}
\usepackage{tabularx}
\usepackage{enumitem}
\usepackage{url}

\title{Formality Style Transfer with Hybrid Textual Annotations}

\author{
Ruochen Xu$^1$
\and
Tao Ge$^2$\and
Furu Wei$^2$
\affiliations
$^1$Carnegie Mellon University\\
$^2$Microsoft Research Asia\\
\emails
ruochenx@cs.cmu.edu,
tage@microsoft.com,
fuwei@microsoft.com
}

\date{}

\begin{document}
\maketitle
\begin{abstract}
Formality style transformation is the task of modifying the formality of a given sentence without changing its content. Its challenge is the lack of large-scale sentence-aligned parallel data. In this paper, we propose an omnivorous model that takes parallel data and formality-classified data jointly to alleviate the data sparsity issue. We empirically demonstrate the effectiveness of our approach by achieving the state-of-art performance on a recently proposed benchmark dataset of formality transfer. Furthermore, our model can be readily adapted to other unsupervised text style transfer tasks like unsupervised sentiment transfer and achieve competitive results on three widely recognized benchmarks.
\end{abstract}

\section{Introduction}
Text style transfer is an important research topic in natural language generation since specific styles of text are preferred in different cases \cite{sennrich2016controlling,rabinovich2017personalized}. Early work of style transfer relies on parallel corpora where paired sentences have the same content but are in different styles. For example, \citeauthor{xu2012paraphrasing} \shortcite{xu2012paraphrasing} studied the task of paraphrasing with a target writing style. Trained on human-annotated sentence-aligned parallel corpora, their model can transfer text into William Shakespeare's style. 

However, the genre and amount of parallel corpora is very limited for text style transfer tasks. Therefore, recent work focuses on eliminating the requirement for parallel corpora \cite{mueller2017sequence,hu2017toward,shen2017style,fu2018style,li2018delete,xu2018unpaired,dos2018fighting,melnyk2017improved,tian2018structured}. A common strategy is to first learn a style-independent representation of the content in the input sentence. The output sentence is then generated based on the content and the desired style.
To this end, some approaches \cite{mueller2017sequence,hu2017toward,shen2017style,fu2018style,melnyk2017improved} leverage auto-encoder frameworks, where the encoder generates a hidden representation of the input sentence. Other work including \cite{xu2018unpaired,li2018delete,zhang2018learning}, on the other hand, explicitly deletes those style-related keywords in order to form style-independent sentences.

\begin{figure}[ht]
\includegraphics[width=0.50\textwidth]{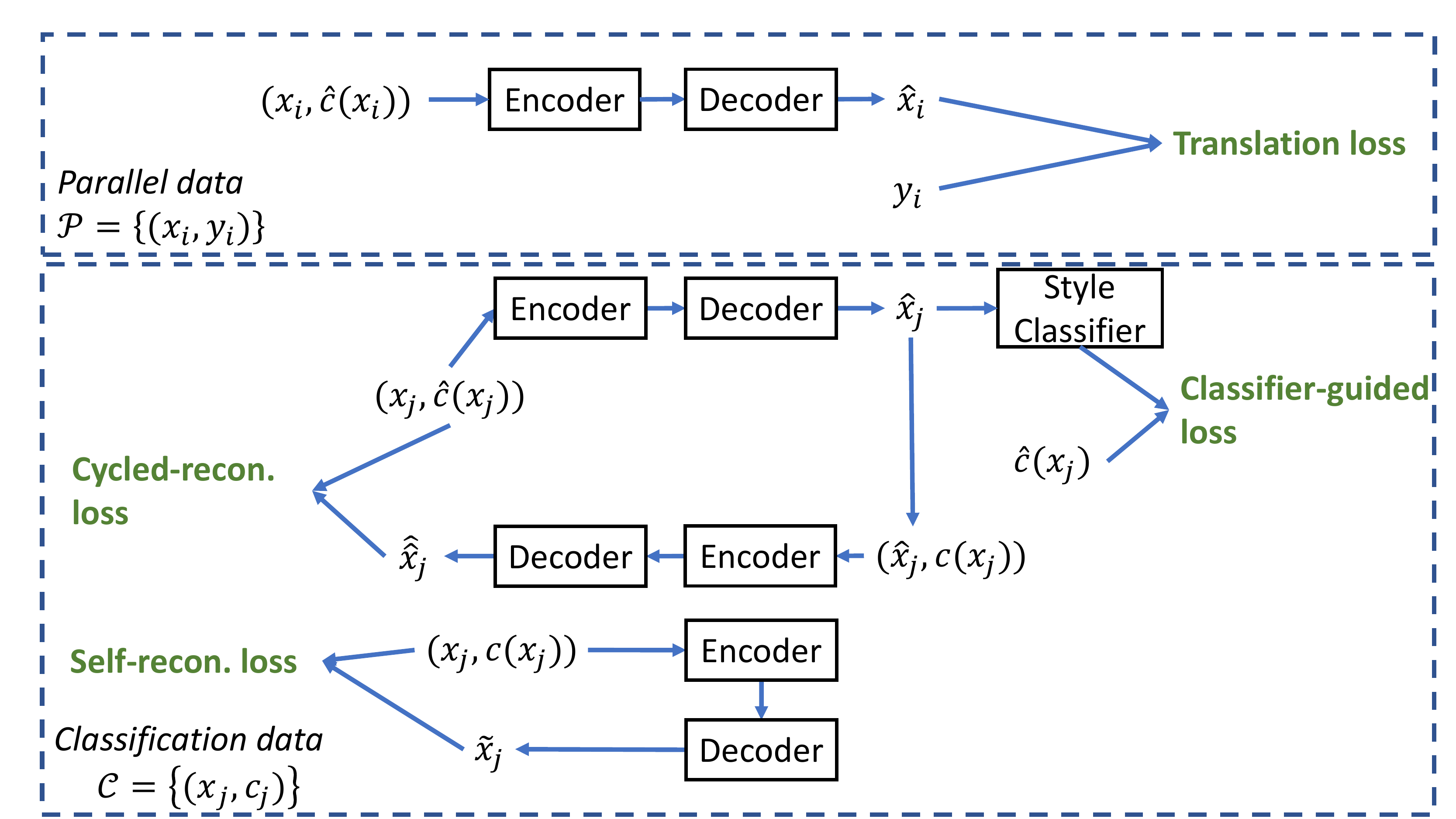}
\centering
\caption{The model architecture and various losses for the formality transformer model. All the encoders(decoders) in the figure refer to the same model which appears repeatedly in different loss functions. We use $\Tilde{x}_j$ to represent $x_j$ after self-reconstruction and $\hat{\hat{x_j}}$ to represent $x_j$ after cycled-reconstruction}
\label{fig:arch}
\end{figure}

The key insight of these unsupervised methods for text style transfer is to disentangle style information from content. To achieve this, an adversarial \cite{fu2018style,shen2017style}, collaborative classifier \cite{melnyk2017improved}, or language model \cite{yang2018unsupervised} is used to guide the generation process. 
Some work \cite{melnyk2017improved,xu2018unpaired} applies reconstruction losses to ensure that the semantic content of the input sentence can be recovered after style transformation.

We take the insights from these unsupervised methods and propose a novel formality transfer model that is able to incorporate hybrid types of textual annotations. As Figure \ref{fig:arch} shows, our approach involves a sequence-to-sequence (seq2seq) encoder-decoder model for style transformation and a style classification model. The proposed seq2seq model simultaneously handles the bidirectional transformation of formality between informal and formal, which not only enhances the model's data efficiency but also enables various reconstruction losses to help model training; while the classification model fully utilizes the less expensive formality-classified annotation to compute a classifier-guided loss as additional feedback to the seq2seq model. Experiments on multiple benchmark datasets demonstrate that our approach not only achieves the best performance on formality style transfer but also can be easily adapted to other text style transfer tasks with competitive performance.  

To summarize, our major contributions include:
\begin{itemize}
    \item We advanced the state-of-art on the formality transfer task, with a novel approach that could be trained on both limited parallel data and larger unpaired data.
    \item We propose a bi-directional text style transfer framework that can transfer formality from formal to informal or from informal to formal with one single encoder-decoder component, enhancing the models' data efficiency. With jointly optimized against various losses, the models can be better trained and yield a promising result.
    \item Our approach could be easily generalized to unsupervised setting and to other text style transfer tasks like sentiment transfer.
\end{itemize}



\section{Background}
Transferring the formality is an important dimension of style transfer in text \cite{heylighen1999formality}, whose goal is to change the style of a given content to make it more formal. The formality transfer system would be useful in addition to any writing assistant tool. To illustrate the desired output of a formality transfer system, we show in table \ref{tab:formal-exp} the examples of formal rewrites of informal sentences from a recent formality transfer dataset \cite{rao2018dear}.

\begin{table}[bt]
\centering
\resizebox{0.5\textwidth}{!}{%
\begin{tabular}{ll}
\hline
Informal & I'd say it is punk though. \\
Formal & However, I do believe it to be punk. \\ \hline
Informal & Gotta see both sides of the story. \\
Formal & You have to consider both sides of the story. \\ \hline
\end{tabular}
}
\caption{Examples from dataset introduced by \protect\cite{rao2018dear}, the formal sentences are the rewrites from the informal ones annotated by human experts.}
\label{tab:formal-exp}
\end{table}

One typical solution to formality transfer could be the seq2seq encoder-decoder framework \cite{bahdanau2014neural}, which has been successful for machine translation and other text-to-text tasks. 
Given an informal sentence $\boldsymbol{x}=(x_1,\cdots,x_M)$ and its corresponding formal rewrite sentence $\boldsymbol{y}=(y_1,\cdots,y_N)$ in which $x_M$ and $y_N$ are the $M$-th and $N$-th words of sentence $\boldsymbol{x}$ and $\boldsymbol{y}$ respectively, the seq2seq model learns a probabilistic mapping $P(\boldsymbol{y}|\boldsymbol{x})$ from the parallel sentence pairs through maximum likelihood estimation (MLE), which learns model parameters $\boldsymbol{\Theta}$ to maximize the following equation: 

\begin{equation}
\small
\label{eq:loss-seq2seq}
\boldsymbol{\Theta^*} = \arg \max_{\boldsymbol{\Theta}} \sum_{(\boldsymbol{x},\boldsymbol{y}) \in \mathcal{P}} \log P(\boldsymbol{y}|\boldsymbol{x};\boldsymbol{\Theta})
\end{equation}
where $\mathcal{P}$ denotes the set of the parallel informal-formal sentence pairs.

\section{Joint Training with Hybrid Textual Annotation}
Notwithstanding the seq2seq model's effectiveness, it requires large amounts of parallel data for training to update its millions of parameters. Unfortunately, the parallel data for formality style transfer is expensive. As a result, there are only a limited number of informal-formal parallel sentence pairs available for training, which hinders training a good seq2seq model with the conventional training method that largely relies on the parallel data. To address the limitation of the conventional seq2seq training, we propose a novel joint training approach that is able to train a seq2seq model jointly from hybrid textual annotations (i.e., from both parallel annotation and class-labeled annotation).


Assume we have two sets of data: $\mathcal{P}$ and $\mathcal{C}$.
$\mathcal{P}$ is the set of parallel sentence pairs: $\mathcal{P} = \{ ( \boldsymbol{x^{(i)}}, \boldsymbol{y^{(i)}}) \}_{i=1}^{\|\mathcal{P}\|}$, where the $i$-th sentence pair contains informal sentence $\boldsymbol{x^{(i)}}$ and its corresponding formal re-writing $\boldsymbol{y^{(i)}}$ and $\boldsymbol{x^{(i)}}$ and $\boldsymbol{y^{(i)}}$ are expected to only differ in terms of their formality of expression while their semantic content must be the same. $\mathcal{C}$ is the classification data: $\mathcal{C} = \{ (\boldsymbol{x^{(j)}}, c^{(j)}) \}_{j=1}^{\|\mathcal{C}\|}$. $\boldsymbol{x^{(j)}}$ is a sentence with the formality label $c^{(j)}$ where $c^{(j)} \in \{informal, formal\}$. Specially, we use $c(\boldsymbol{x})$ to represent the known formality label for a given sentence $\boldsymbol{x}$, and $\hat{c}(\boldsymbol{x})$ to represent to the opposite formality of $c(\boldsymbol{x})$.

We propose a novel approach to fully exploit $\mathcal{P}$ and $\mathcal{C}$ by jointly training a seq2seq style transformation model parameterized by $\boldsymbol{\Theta_{s2s}}$ and a classifier parameterized by $\boldsymbol{\Theta_c}$ to minimize the losses described in the sub-sections below.

\subsection{Bidirectional Translation Loss}
As most text-to-text tasks, the most direct way to train a model is to minimize the translation loss as defined in Eq (\ref{eq:loss-seq2seq}). In contrast to the conventional seq2seq model that transfers style in only one direction, we propose to model bidirectional style transformation  (i.e., both from informal to formal and from formal to informal) with one single encoder-decoder component. We will show in the following sections that the bidirectional style transformation modeling can not only make full use of the parallel data but also enable various reconstruction constraints to help the models learn better from massive monolingual data, which enhances the models' data efficiency. Different from the conventional seq2seq model where only a source sentence is fed into the model, we also tell the model a direction indicator.  A special token ``$<$to\_formal$>$''(or ``$<$to\_informal$>$'') is appended at the beginning of the input sentence and fed into the encoder in order to specify the direction of the transfer.

For modeling bidirectional style transformation, for each paired sentence $(\boldsymbol{x}, \boldsymbol{y}) \in \mathcal{P}$, we minimize the negative log likelihood of generating $\boldsymbol{y}$ given $\boldsymbol{x}$ and the reverse direction.

\begin{align}
    \label{eq:loss-trans}
     L_{trans}(\boldsymbol{\Theta_{s2s}}) = & \nonumber \\
     -\sum_{(\boldsymbol{x}, \boldsymbol{y}) \in \mathcal{P}} & \log P(\boldsymbol{y} | \boldsymbol{x}, \hat{c}(\boldsymbol{x}); \boldsymbol{\Theta_{s2s}}) \nonumber  + \nonumber \\
    & \log P(\boldsymbol{x} | \boldsymbol{y}, \hat{c}(\boldsymbol{y}); \boldsymbol{\Theta_{s2s}})
\end{align}

As shown in Eq (\ref{eq:loss-trans}), the translation loss is defined on both directions of a sentence pair in parallel data. With shared parameters for bi-directional translation, the model can be trained from parallel sentence pairs in both directions, making the size of training data twice and accordingly improving the model's data efficiency.

However, the size of parallel data $\|\mathcal{P}\|$ is still too small to train a well-generalized seq2seq model for formality transfer. To avoid the overfitting problem, we further introduce the following losses.

\subsection{Classifier-guided Loss}
To assist model training from the limited parallel data, we propose to use the classification data $\mathcal{C}$ which is much more easily accessible than the parallel data. We first train a classifier to predict the formality of a given sentence. The objective for training the formality classifier is the standard negative log-likelihood loss given in equation \ref{eq:loss-pre-clas}

\begin{equation}
    \label{eq:loss-pre-clas}
    L_{clas}(\boldsymbol{\Theta_c}) =  -\sum_{(\boldsymbol{x}, c(x)) \in \mathcal{C}} \log P(c(x)|\boldsymbol{x}; \boldsymbol{\Theta_c})
\end{equation}

After the classifier $\boldsymbol{\Theta_c}$ is learned, we keep its parameters fixed and use it to update the seq2seq model. Given an informal sentence $\boldsymbol{x}$ and the desired formality $\hat{c}(x)$ (formal), we let $Seq2Seq(\boldsymbol{x}, \hat{c}(\boldsymbol{x}))$ be the transferred sentence given by the seq2seq model($\boldsymbol{\Theta_{s2s}}$). In other words, we assume we can find $$Seq2Seq(\boldsymbol{x}, \hat{c}(\boldsymbol{x}) ) = \arg\max_{y} P(y | \boldsymbol{x}, \hat{c}(\boldsymbol{x}); \boldsymbol{\Theta_{s2s}} )$$

Since in classification data $\mathcal{C}$, we do not have the ground truth of $Seq2Seq(\boldsymbol{x}, \hat{c}(\boldsymbol{x}))$, we cannot optimize $\boldsymbol{\Theta_{s2s}}$ as in Eq \ref{eq:loss-trans}. Alternatively, we can optimize classification loss given by the trained formality classifier in order to let $Seq2Seq(\boldsymbol{x}, \hat{c}(\boldsymbol{x}))$ look like the desired style $\hat{c}(\boldsymbol{x})$. The loss is shown in Eq (\ref{eq:loss-clas}):


\begin{align}
    \label{eq:loss-clas}
    & L_{clas-guided}(\boldsymbol{\Theta_{s2s}}) = \nonumber \\
    & -\sum_{(\boldsymbol{x}, c(x)) \in \mathcal{C}} \log P(\hat{c}(\boldsymbol{x})|Seq2Seq(\boldsymbol{x}, \hat{c}(\boldsymbol{x})); \boldsymbol{\Theta_c})
\end{align}

\subsubsection{Differentiable Decoding}

In order to generate $Seq2Seq(x, c)$, the decoder samples the output sequence $y_1, y_2, ..., y_{L_y}$ of tokens one element at a time. The process is auto-regressive in the sense that previously generated tokens are fed into the decoder to generate the next token.
In its original formulation, the classifier-guided loss (equation \ref{eq:loss-clas}) contains discrete samples generated in such auto-regressive manner, which hinders the gradient propagation. To solve this, we apply a recent technique \cite{hu2017toward,shen2017style} to approximate the discrete decoding process with a differentiable one. Instead of feeding a discretely sampled token to the decoder, we feed the softmax distribution over the vocabulary as the generated soft word. Let the output logit vector at time step $t$ be $v_t$. The output for the decoder is $softmax(v_t/\tau)$, where $\tau \in (0, 1)$ is the temperature hyper-parameter to control the shape of the softmax function. For the embedding look-up layer of the decoder and the classifier, we simply take ``soft'' word embedding by averaging over the word embedding matrix. The generated soft embedding is differentiable w.r.t. the parameters in the decoder and encoder, which enables the gradient from the classifier to propagate back and update the formality transfer model.

\subsection{Reconstruction Loss}\label{subsec:reconstruction}
One potential issue of using classifier-guided loss is that the seq2seq model could easily minimize the classification loss defined in Eq (\ref{eq:loss-clas}) by simply generating keywords for $\hat{c}(\boldsymbol{x})$. In this case $Seq2Seq(\boldsymbol{x}, \hat{c}(\boldsymbol{x}))$ will be classified to the target formality class but become independent to input $\boldsymbol{x}$. To overcome this problem, we propose two reconstruction losses that are easily introduced based on our bi-directional transformation modeling framework.

The first loss is the self-reconstruction loss, which encourages the seq2seq model to reconstruct the input itself if the desired formality is the same as the input one. The objective is similar to Eq (\ref{eq:loss-trans}) and is defined using the maximum likelihood as in Eq (\ref{eq:loss-self-recon}).

\begin{align}
    \label{eq:loss-self-recon}
    & L_{self-recon}(\boldsymbol{\Theta_{s2s}})  = \nonumber \\
    & -\sum_{(\boldsymbol{x}, c(x)) \in \mathcal{C}} \log P(\boldsymbol{x}) | \boldsymbol{x}, c(\boldsymbol{x}); \boldsymbol{\Theta_{s2s}})
\end{align}

In other words, the self-reconstruction loss makes the seq2seq model leave the input sentence untouched if the desired formality already exists in the input.

The second reconstruction loss requires the seq2seq model's ability to reconstruct the input sentence after a looped transformation which first transfers the input sentence to the target formality and then transfers the output back to its original formality. Let $\hat{\boldsymbol{x}} = Seq2Seq(\boldsymbol{x}, \hat{c}(\boldsymbol{x}))$, the cycled-reconstruction loss is defined in Eq (\ref{eq:loss-cyc-recon}):

\begin{equation}
    \label{eq:loss-cyc-recon}
    L_{cyc-recon}(\boldsymbol{\Theta_{s2s}}) = 
     -\sum_{(\boldsymbol{x}, c(x)) \in \mathcal{C}} \log P(\boldsymbol{x} | \hat{\boldsymbol{x}}, c(\boldsymbol{x}); \boldsymbol{\Theta_{s2s}})
\end{equation}

\subsubsection{One-sided Cycle Approximation}
The discrete generation also exists in the cycled reconstruction loss in Eq (\ref{eq:loss-cyc-recon}). Instead of using the differentiable decoding, we take a simpler approach by only back-propagating the gradient until the generation of $Seq2Seq(\boldsymbol{x}, \hat{c}(\boldsymbol{x}))$. In other words, we take $(Seq2Seq(\boldsymbol{x}, \hat{c}(\boldsymbol{x})), \boldsymbol{x})$ as a pair of pseudo-parallel data. The approximation has the same form of back-translation as used in machine translation \cite{sennrich2015improving}. The difference is that our model works in monolingual data and that we have a formality label $c$ to control the direction of transformation, whereas in machine translation the back-translation needs a separate back-translator of the reverse direction.

\subsection{Overall Objective}

The overall objective of our seq2seq model is the weighted summation of the various losses we defined above, except for the classification loss in Eq (\ref{eq:loss-overall}), which is defined on formality classifier and is optimized as a separate step.

\begin{align}
    \label{eq:loss-overall}
    L_{all}(\boldsymbol{\Theta_{s2s}}) = & w_t L_{trans} + w_c L_{clas-guided} \nonumber \\
    + & w_{sr} L_{self-recon} + w_{cr} L_{cyc-recon}
\end{align}
\section{Model Configuration}
In this section, we illustrate the detailed configuration of the seq2seq model and the classification model in our approach as well as our post-processing steps for formality style transfer.

\subsection{Formality Transformer Model}
For the seq2seq model, we use the recently proposed transformer model \cite{vaswani2017attention}. In contrast to conventional RNN seq2seq models, a transformer model applies self-attention to the source sentence, which intuitively benefits disambiguation of word sense in an informal sentence and should accordingly yield a better style transfer result. To the best of our knowledge, this is the first attempt to adapt the transformer model on formality transformation.

We implement the transformer model based on open source sequence-to-sequence software Fairseq-py \cite{gehring2017convolutional}
For the transformer model, we use the same configuration for both the encoder and decoder. We set the embedding dimension to $256$ and the hidden dimension of the feed-forward sub-layer to $1024$. The number of layers is set to $2$ and the number of heads is set to $4$. For the remaining hyper-parameters such as the dropout rate and the activation function, we followed the default choice of Fairseq-py in our implementation. For temperature $\tau$, we anneal it from $1.0$ to $0.1$ as training proceeds.

For hyper-parameter tunning, we performed grid search over intervals $[0.1, 0,2, 0.5, 1.0]$ for each weight in equation \ref{eq:loss-overall} in the development set of GYAFC dataset(see section \ref{subsec:exp}).

\subsection{Formality Classification Model}
For our formality classification model, we use a CNN text classifier \cite{kim2014convolutional}. Given an input sentence $x$, we first embedded each token with word embedding layer. Then convolutional filters of size $n$ is applied on the embedded sentence, which acts as $n$-gram feature extractors on the different position of the input sentence. We then take the maximum of the extracted features over different positions with a max-pooling layer. The final feed-forward layer has softmax activation to produce the probability of a given formality.

For implementation details, we use filter size $n=\{3,4,5\}$ and filter number $100$ for each size. The dropout rate is set to $0.5$. 

\subsection{Post-processing}\label{subsec:postprocessing}
\subsubsection{Classifier-based Filtering} \label{subsec:class-filter}
We filter some of the n-best sentences with our formality classifier. The sentences with the incorrect formality class given by the classifier are removed from the candidate list. And the remaining one with the highest generation score is selected as the final output.

\subsubsection{Grammatical Error Correction}
Motivated by the fact that a formal sentence should be grammatically correct, we feed the output from our formality transfer system to a grammatical error correction (GEC) model as post-processing to get rid of the grammatical mistakes. Specifically, we used the state-of-the-art GEC system \cite{ge2018reaching} which is based on a convolutional seq2seq model with special training and inference mechanism to correct grammatical errors in target sentences.

\section{Experiment}


\subsection{Experimental Setting}
\label{subsec:exp}
\subsubsection{Data}
We used Grammarly's Yahoo Answers Formality Corpus (GYAFC) \cite{rao2018dear} as our evaluation dataset. It contains paired informal and formal sentences annotated by humans. The dataset is crawled and annotated from two domains in Yahoo Answers \footnote{\url{https://answers.yahoo.com}}, namely Entertainment \& Music (E\&M) and Family \& Relationships (F\&R) categories. Following the analysis in \cite{rao2018dear}, we only consider the transformation from informal to formal. The statistics of the training, validation and testing splits of GYAFC dataset is shown in table \ref{tab:GYAFC-stat}.

\begin{table}[ht]
\centering
\begin{tabular}{@{}llll@{}}
\toprule
 & Train & Validate & Test \\ \midrule
E\&M & 52595 & 2877 & 1416 \\
F\&R & 51967 & 2788 & 1332 \\ \bottomrule
\end{tabular}
\caption{The statistics of train, validate and test set of GYAFC.
}
\label{tab:GYAFC-stat}
\end{table}

All splits in GYAFC are given in the form of parallel data. To obtain the classification data, we apply a simple data extension heuristic. We first train a CNN formality classifier on the training split of the parallel corpora, and then make predictions on the unlabeled corpus of Yahoo Answers. The predictions of either formal or informal with a confidence higher than 99.5\% are selected as the classification dataset $\mathcal{C}$, which contains about 1100k formal sentences and 350k informal sentences. During training, our model combines the training data from E\&M and F\&R. The relative weights for various losses are tuned on the validation set.

\subsubsection{Baselines}
We compare to the following baseline approaches:
\begin{itemize}[leftmargin=*]
    \setlength\itemsep{0em}
    \item \textbf{Transformer} is the original transformer model \cite{vaswani2017attention} that shares the same configurations of our model.
    \item \textbf{Transformer-Combine} is \textbf{Transformer} that combines the training data from E\&M and F\&R.
    \item \textbf{SimpleCopy} is to simply copy the input sentence as the prediction without any modification.
    \item \textbf{RuleBased} is the rule-based method that uses a set of rules to automatically make an informal sentence more formal.
    \item \textbf{PBMT} is a phrase-based machine translation model trained on parallel training data and outputs from \textbf{RuleBased}.
    \item \textbf{NMT} is the encoder-decoder model with attention mechanism \cite{bahdanau2014neural}.
    \item \textbf{NMT-Copy} adds the copy mechanism \cite{gu2016incorporating} to \textbf{NMT}.
    \item \textbf{NMT-PBMT} is a semi-supervised method that further incorporates outputs from \textbf{PBMT} with back-translation. Both domain knowledge and additional unlabeled data were used to train this strong baseline.
    \item \textbf{MultiTask} \cite{niu2018multi} jointly trains the model on GYAFC and a formality-sensitive machine translation task with additional bilingual supervision. It also uses ensemble decoding, while our model and other baselines all use single model decoding. Therefore the performance is not comparable with other models. 
    
\end{itemize}
Note that in addition to the first three baselines that we implement by ourselves, the outputs of the remaining baselines are from previous works \cite{rao2018dear,niu2018multi}.

\subsubsection{Evaluation Metric}
We followed the evaluation metric used in \cite{rao2018dear,li2018delete} to use BLEU \cite{papineni2002bleu} to measure the closeness between the system prediction and the human annotation. Moreover, we use GLEU score \cite{napoles2015ground} as an alternative to BLEU. GLEU was originally introduced in the task of grammatical error correction (GEC), which is generalized and modified from BLEU to address monolingual text rewriting evaluation and shows better correlation with human evaluations than BLEU. 
\subsection{Results and Analysis}

\begin{table}[ht]
\centering
\resizebox{0.5\textwidth}{!}{%
\begin{tabular}{lll|ll}
\hline
 & \multicolumn{2}{l|}{E\&M} & \multicolumn{2}{l}{F\&R} \\ \hline
 & BLEU & GLEU & BLEU & GLEU \\ \hline
RuleBased & 60.37 & 16.48 & 66.4 & 18.79 \\
PBMT & 66.88 & 24.38 & 72.4 & 26.96 \\
NMT & 58.27 & 22.87 & 68.26 & 26.3 \\
NMT-Copy & 58.67 & 22.93 & 68.09 & 26.05 \\
NMT-Combine & 67.51 & 24.05 & 73.78 & 26.74 \\
SimpleCopy & 50.28 & 7.42 & 51.66 & 6.8 \\
Transformer & 61.86 & 21.61 & 66.69 & 24.94 \\
Transformer-Combine & 65.5 & 23.94 & 70.63 & 25.88 \\ \hline
MultiTask* & 72.01 & 25.92 & 75.35 & 27.15 \\ \hline
Ablt. w/o self-recon & 64.53 & 22.81 & 70.43 & 22.92 \\
Ablt. w/o cyc-recon & 66.39 & 23.53 & 71.71 & 25.50 \\
Ablt. w/o class-guided & 67.90 & 24.13 & 72.00 & 24.64 \\ \hline
Ours & 69.08 & 24.37 & 72.90 & 24.78 \\
Ours w/ class-filter & 68.71 & 24.64 & 73.16 & 25.73 \\
Ours w/ gec & \textbf{69.63} & \textbf{25.78} & \textbf{74.43} & \textbf{27.35} \\ \hline
\end{tabular}
}
\caption{BLEU and GLEU scores on GYAFC dataset. The dataset has two domains: Entertainment \& Music (E\&M) and Family \& Relationship (F\&R). The best single model score under each metric is marked bold. \textbf{MultiTask*} is not comparable to other models in the table since it uses more supervised data and ensemble decoding.}
\label{tab:GYAFC-res}
\end{table}

The results for formality transfer on GYAFC dataset is shown in table \ref{tab:GYAFC-res}. \textbf{Ours} represents our approach's results without post-processing, while \textbf{Ours w/ class-filter} and \textbf{Ours w/ gec} are the results with the post-processing steps introduced in Section \ref{subsec:postprocessing}.

According to Table \ref{tab:GYAFC-res}, there are several noteworthy points: Firstly, it is observed that the best single-model performance is \textbf{Ours w/ gec}, which outperformed all baseline methods in terms of both BLEU and GLEU in E\&M and F\&R. Compared to the strongest baselines PBMT and NMT-PBMT, both of which incorporate heuristic rules with the training data, our models are end-to-end neural networks without any prior knowledge of the task. The performance of our single model is also competitive with the state-of-art ensemble model(MultiTask), which also utilizes additional supervision.

Secondly, our model \textbf{Ours} outperformed Transformer-Combine by a large margin for both BLEU and GLEU and on both domains. We can conclude that our joint training with hybrid annotation (parallel data and classification data) could significantly improve the performance of the state-of-art seq2seq model on the formality transformation task. 

\begin{table*}[ht]
\centering
\begin{adjustbox}{max width=\textwidth}
\begin{tabular}{lllll|llll|llll}
\hline
 & \multicolumn{4}{l}{Yelp} & \multicolumn{4}{l|}{Amazon} & \multicolumn{4}{l|}{ImageCaption} \\ \hline
 & ACC & BLEU & G-score & GLEU & ACC & BLEU & G-score & GLEU & ACC & BLEU & G-score & GLEU \\ \hline
CrossAligned & 73.7 & 8.12 & 24.46 & 5.61 & 74.3 & 1.81 & 11.60 & 1.54 & 74.1 & 1.82 & 11.61 & 1.57 \\
StyleEmbedding & 8.7 & 19.50 & 13.02 & 7.19 & 54.7 & 14.60 & 28.26 & 8.17 & 43.3 & 8.76 & 19.48 & 5.93 \\
MultiDecoder & 47.6 & 13.25 & 25.11 & 6.37 & 68.5 & 8.72 & 24.44 & 5.40 & 68.3 & 6.63 & 21.28 & 4.56 \\
TemplateBased & 81.7 & 21.05 & 41.47 & 11.43 & 92.5 & 34.18 & \textbf{56.23} & 19.07 & 68.7 & 19.10 & 36.22 & \textbf{12.65} \\
RetrieveOnly & \textbf{95.4} & 1.52 & 12.04 & 1.31 & 95.5 & 2.61 & 15.79 & 2.18 & 70.3 & 2.66 & 13.67 & 2.06 \\
DeleteOnly & 85.7 & 13.59 & 34.13 & 8.33 & 83.0 & - & - & 7.29 & 45.6 & 11.91 & 23.30 & 7.81 \\
DeleteAndRetrieve & 88.7 & 14.75 & 36.17 & 8.69 & \textbf{96.8} & 29.59 & 53.52 & 17.38 & 48.0 & 11.94 & 23.94 & 7.86 \\
SimpleCopy & 3.00 & \textbf{29.64} & 9.43 & 7.58 & 17.80 & \textbf{48.63} & 29.42 & 20.73 & 50.00 & \textbf{19.18} & 30.97 & 11.30 \\ \hline
Ours & 45.80 & 28.57 & 36.17 & 13.73 & 23.00 & 46.31 & 32.64 & \textbf{22.64} & 50.66 & 19.01 & 31.03 & 11.20 \\
Ours w/ class-filter & 82.80 & 26.44 & \textbf{46.79} & \textbf{14.60} & 76.30 & 34.34 & 51.19 & 18.84 & \textbf{79.66} & 17.01 & \textbf{36.81} & 10.20 \\ \hline
\end{tabular}
\end{adjustbox}
\caption{The performance for unsupervised style transfer dataset. G-score is the geometric mean of Accuracy and BLEU. For accuracy (ACC) of baselines from \protect\cite{li2018delete}, we use their reported numbers. The BLEU and GLEU are evaluated with our own script. The "-" in the table is due to the fact that the provided output misaligned with the ground truth.}
\label{tab:unsup-res}
\end{table*}

Thirdly, we notice that using classifier-filtered decoding (\textbf{Ours w/ class-filter}) outperformed using beam search (\textbf{Ours}) with regard to GLEU for both domains and with regard to BLEU for F\&R. The relative improvement showed that the formality classifier is helpful not only during the training of seq2seq model but also in the testing phase. Also, we verified that the usage of the GEC system as a post-processing step could further improve the performance of the informal to formal transformation.

Lastly, comparing with \textbf{Transformer} and \textbf{Transformer-Combine}, we can see that using the training data from both domains could further improve the performance for both BLEU and GLEU, indicating that with a limited parallel corpus, additional annotation from a different domain could still help with the sequence-to-sequence learning.



\subsection{Ablation Study}
In table \ref{tab:GYAFC-res}, we include the performance of ablated models (\textbf{Ablt. w/o ...}) which exclude one specific loss in the overall objective function. From the results of our ablated models, it is clear that all the losses defined on classification data contribute to the improvement. Another observation is that the performance of \textbf{Ablt. w/o self-recon} is even lower than \textbf{Transformer-Combine}, which demonstrates the importance of the self-reconstruction loss to prevent classifier-guided loss from drastically changing the content of the input sentence. 

\subsection{Case Study}
We sampled some outputs of our model (\textbf{Ours w/ gec}) in the test set and qualitatively compare them with the ground truth in the appendix.



\begin{table}[ht]
\centering
\begin{tabular}{lllll}
\hline
Source & Attribute & Train & Validate & Test \\ \hline
\multirow{2}{*}{Yelp} & Positive & 270K & 2000 & 500 \\
 & Negative & 180K & 2000 & 500 \\ \hline
\multirow{2}{*}{Amazon} & Positive & 277K & 985 & 500 \\
 & Negative & 278K & 1015 & 500 \\ \hline
ImageCaption & Romantic & 6000 & 300 & 0 \\
 & Humorous & 6000 & 6000 & 0 \\
 & Factual & 0 & 0 & 300 \\ \hline
\end{tabular}
\caption{The statistics of sentiment transfer datasets from Yelp and Amazon}
\label{tab:Senti-stat}
\end{table}

\subsection{Experiments on Unsupervised Style Transfer}
In addition to GYAFC, we also verified our framework on three other style transfer datasets from \cite{li2018delete}. The datasets are adapted from sentiment classification and image caption tasks which only contain classification data $\mathcal{C}$. 
The statistics of the three style transfer datasets is shown in the table \ref{tab:Senti-stat}.
The goal for Yelp and Amazon is to transfer the sentiment of a online review either from negative to positive or vice-versa, while the goal of ImageCaption is to transfer between romantic and humorous image captions without the image. Note that ImageCaption has only textual testing sentences, which are supposed to contain no style, as testing data. We follow the procedure in \cite{li2018delete} and treat them as the source style sentences without any additional post-processing. Due to lack of parallel corpus, our model only has unsupervised objective excluding the translation loss (equation \ref{eq:loss-trans}). We set loss weights as $w_c=1.0$, $w_{sr}=0.5$ and $w_{cr}=1.0$ for all our models.

\subsubsection{Baseline Methods}
For the unsupervised sentiment transfer task, we compare our unsupervised method with the baselines outputs from \cite{li2018delete}.

\begin{itemize}[leftmargin=*]
    \setlength\itemsep{0em}
    \item \textbf{RetrieveOnly} returns a retrieved sentence from the corpus of the target sentiment, using the source sentence as the query.
    \item \textbf{TemplateBased} finds sentiment keywords by the statistics from the classification dataset. It then replaces the keywords in the source sentence with the ones in the \textbf{RetrieveOnly}.
    \item \textbf{DeleteOnly} \cite{li2018delete} learns a RNN-based seq2seq model. The objective is to reconstruct the training source sentence with the sentiment keywords being removed.
    \item \textbf{DeleteAndRetrieve} \cite{li2018delete} is similar to \textbf{DeleteOnly} but further guides the replacement of the sentiment keywords with the retrieved sentence.
    \item \textbf{StyleEmbedding} \cite{fu2018style} also uses seq2seq model, where the encoder tries to learn a style-independent vector representation of the input sentence, from which an adversarial discriminator cannot tell the style of the input. 
    \item \textbf{MultiDecoder} \cite{fu2018style} is similar to \textbf{StyleEmbedding}, except that it uses multiple decoders for different styles.
    \item \textbf{CrossAligned} \cite{shen2017style} instead uses the adversarial discriminator on the hidden states of the recurrent neural network (RNN) decoder.    
\end{itemize}


\subsubsection{Evaluation Metric}
Following \cite{li2018delete}, we report the accuracy and BLEU of the transferred sentence in the unsupervised tasks. The accuracy is the percentage of the generated sentences that contains the desired style, which is determined by a pre-trained style classifier. \footnote{We used a separate classifier other than the one used in training.} We notice that there is a trade-off between the style accuracy and BLEU point for all three datasets. To evaluate the overall performance, we follow \cite{xu2018unpaired} and use the geometric mean of accuracy and BLEU as an evaluation metric: G-score. Intuitively, methods that make aggressive modifications will achieve higher accuracy but suffer from poor semantic content preserving and thus low BLEU.


\subsubsection{Results}

The results for sentiment transfer on Yelp and Amazon dataset is shown in table \ref{tab:unsup-res}.




Our model achieves the highest G-score in Yelp and ImageCaption dataset, and the highest GLEU in Yelp and Amazon dataset. The promising results from various metrics suggest that our methods could effectively modify the style-dependent span of text accordingly while keeping the content unchanged.

\section{Conclusion}
In this paper, we present an approach to training formality transfer models from hybrid textual annotations. Based on a bidirectional style transformation seq2seq model, we fully exploit formality style classification data through classification feedback and various reconstruction constraints to assist the model learning. Our approach effectively improved the performance of the base model and achieved new state-of-art results on formality transfer task. Furthermore, our approach can be readily generalized to other unsupervised style transfer tasks and perform consistently well on multiple benchmarks.

\bibliographystyle{named}
\bibliography{ref}

\appendix

\section{Case Study}

\begin{table*}[h]
\centering
\resizebox{1.0\textwidth}{!}{%
\begin{tabular}{ll}
\hline
Case 1 & Fail to recognize named entities \\ \hline
source & secondly id pick yellow card then simple plan then bowling for soup and finally red hot chilli peppers . \\
hypothesis & I would pick yellow card , then try bowling for soup and finally red hot chilli peppers . \\
ground truth & I would pick Yellow Card second , then Simple Plan , Bowling for Soup , and finally , Red Hot Chili Peppers . \\ \hline
Case 2 & Only transfer part of the input \\ \hline
source & I think its a bit cheezy to name it though , but maybe I maybe name it Jet \\
hypothesis & I think it is a bit cheezy to name it . \\
ground truth & It 's a cheesy name but you could name it Jet . \\ \hline
Case 3 & Fail to transfer informal words and phrases \\ \hline
source & Not at all , I really dont think in Spanish either ... they rap about stupid stuff \\
hypothesis & Not at all . I do not think in Spanish either . They rap about stupid stuff . \\
ground truth & Not at all . I really do not think it is so in Spanish , either - they rap about ridiculous things . \\ \hline
\end{tabular}
}
\caption{Some typical failure cases for \textbf{Ours w/ gec} on E\&M of GYAFC}
\label{tab:case-study}
\end{table*}

\begin{table*}[bt]
\centering
\begin{tabular}{ll}
\hline
Case 1 & Common informal word replacement \\ \hline
source & Wat did the beaver say 2 the log . \\
hypothesis & What did the beaver say to the log ? \\
ground truth & What did the beaver say to the log ? \\ \hline
Case 2 & Capitalization \\ \hline
source & and how good of a president do u think he would be ? \\
hypothesis & How good of a president do you think he would be ? \\
ground truth & How good of a president do you think he would be ? \\ \hline
Case 3 & Punctuation \\ \hline
source & ( i know how to read tabs ) \\
hypothesis & I know how to read tabs . \\
ground truth & I know how to read tabs . \\ \hline
\end{tabular}
\caption{Some typical successful cases for \textbf{Ours w/ gec} on E\&M of GYAFC}
\label{tab:case-study-good}
\end{table*}

The most common error case is the failure to handle named entities. (case 1 in table \ref{tab:case-study})
Our model fails to recognize them and convert them into upper case. This could also explain why the performance of Entertainment \& Music is lower than Family \& Relationship, since the former domain contains more named entities.

Incomplete transfer (case 2) and neglect of informal words (case 3) are also frequent among failure cases. Similar to the observation in \cite{rao2018dear}, we notice that our model produced similar errors more often when the input sentence is very informal, not fluent, grammatically incorrect or contains unknown/misspelled tokens.

On the other hand, we also analyzed the most accurate predictions of our model. Some of the most common cases are summarized in table \ref{tab:case-study-good}. We notice that our model is particularly good at handling common and typical informal words (case 1) such as "Wat"(What) and "2"(to). Our model is also able to properly capitalize the first letter of the sentence (case 2) and correctly use the punctuations.

\end{document}